\documentclass{llncs}
\usepackage{times}
\usepackage{latexsym}

\usepackage{url}
\usepackage{amsmath}
\usepackage{amssymb}
\usepackage{graphicx}
\usepackage{hhline}
\usepackage{subfigure}

\newcommand{\KL}{\mathop{\rm KL}\nolimits}

\title{Interpretable probabilistic embeddings: bridging the gap between topic models and neural networks}

\author{Anna Potapenko\inst{1} \and Artem Popov\inst{2} \and Konstantin Vorontsov\inst{3}}
\institute{National Research University Higher School of Economics \email{anna.a.potapenko@gmail.com} \and
 Lomonosov Moscow State University \\ \email{popov.artem.s@yandex.ru} \and
 Moscow Institute of Physics and Technology \\
 \email{vokov@forecsys.ru}}

\date{}

\begin{document}
\maketitle

\begin{abstract}  
 We consider probabilistic topic models and more recent word embedding techniques from a perspective of learning hidden semantic representations. Inspired by a striking similarity of the two approaches, we merge them and learn probabilistic embeddings with online EM-algorithm on word co-occurrence data. The resulting embeddings perform on par with Skip-Gram Negative Sampling (SGNS) on word similarity tasks and benefit in the interpretability of the components. Next, we learn probabilistic document embeddings that outperform paragraph2vec on a document similarity task and require less memory and time for training. Finally, we employ multimodal Additive Regularization of Topic Models (ARTM) to obtain a high sparsity and learn embeddings for other modalities, such as timestamps and categories. We observe further improvement of word similarity performance and meaningful inter-modality similarities. 
\end{abstract}

\section{Introduction}

Recent progress in deep natural language understanding prompted a variety of word embedding techniques that work remarkably well for capturing semantics. These techniques are usually considered as general neural networks that predict context words given an input word~\cite{Bengio:2003,Mikolov:2013,Le:2014}. Although this perspective is convenient to generalize to more complex neural network architectures, e.g. skip-thought vectors~\cite{Kiros:2015}, we believe that it is also important to establish connections between neural embeddings and more traditional models of distributional semantics. It gives theoretical insights about certain models and enables to use previous work as a grounding for further advances. 

One of the first findings in this line of research is interpreting Skip-Gram Negative Sampling (SGNS, \cite{Mikolov:2013}) as an implicit matrix factorization of the shifted Pointwise Mutual Information (PMI) matrix~\cite{Levy:2014b}. It brings SGNS to the context of various vector space models (VSMs) developed during the last decades. Pantel and Turney~\cite{Pantel:2010} provide a thorough survey of VSMs dividing them into word-word, word-context and word-document categories based on the type of the co-occurrence matrix. According to the distributional hypothesis~\cite{Harris:1954}, similar words tend to occur in similar contexts; thus the rows of any of these matrices can be used for estimating 
word similarities~\cite{Deerwester:1990}. Gentner~\cite{Gentner:1983} defines attributional similarity (e.g. \textit{dog} and \textit{wolf}) and relational similarity (e.g. \textit{dog:bark} and \textit {cat:meow}), which are referred to as similarity and analogy tasks in more recent papers. While Baroni et al.~\cite{Baroni:2014} argue that word embeddings inspired by neural networks significantly outperform more traditional count-based approaches for both tasks, Levy et al.~\cite{Levy:2015} tune a shared set of hyperparameters and show that two paradigms give a comparable quality. 

We follow this line of research and demonstrate how principle ideas of the modern word embedding techniques and probabilistic topic models can be mutually exchanged to take the best of the two worlds. So far, topic modeling has been widely applied to factorize word-document matrices and reveal hidden topics of document collections~\cite{Hofmann:1999,Blei:2003}. In this paper we apply topic modeling to a word-word matrix to represent words by probabilistic topic distributions. Firstly, we discover a number of practical learning tricks to make the proposed model perform on par with SGNS on word similarity tasks. Secondly, we show that the obtained probabilistic word embeddings (PWE) inherit a number of benefits from topic modeling. 

One such benefit is interpretability. Interpretability of each component as a coherent topic is vital for many downstream NLP tasks. To give an example, exploratory search aims not only to serve similar documents by short or long queries, but also to navigate a user through the results. If a model can explain why certain items are relevant to the query in terms of distinct topics, then these topics can be used to arrange the results by categories.  Murphy et al.~\cite{Murphy:2012} motivated the importance of interpretability and sparsity from the cognitive plausibility perspective and introduced Non-Negative Sparse Embeddings (NNSE), which is a variation of Non-Negative Sparse Coding matrix factorization. 
State-of-the-art techniques, such as 
SGNS or GloVe~\cite{Pennington:2014}
lack both sparsity and interpretability. 
To address this problem, more recent models~\cite{Luo:2016,Sun:2016} extend SGNS and CBOW~\cite{Mikolov:2013} respectively. However, they do that with explicit modifications of optimization procedure, such as project gradient for SGD. A benefit of topic modeling framework is that interpretability comes naturally with a probabilistic interpretation of parameters.

Furthermore, probabilistic word embeddings can be easily extended with Additive Regularization of Topic Models, ARTM~\cite{Vorontsov:2015a}. This is a general framework to combine multiple requirements in one topic model. In this work we use ARTM to obtain sparsity and to learn embeddings for additional \textit{modalities}, such as timestamps, authors, categories, etc. It enables us to investigate inter-modality similarities, because all the embeddings are in the same space. Interestingly, additional modalities also improve performance on word similarity task.
Finally, we build probabilistic document embeddings and show that they outperform DBOW architecture of paragraph2vec~\cite{Le:2014} on a document similarity task.
Thus, we get a powerful framework for learning probabilistic embeddings for various items and with various requirements. 
We train these models with online EM-algorithm similar to~\cite{Hoffman:2010} in BigARTM open-source library~\cite{Frei:2015}.


Related work includes Word Network Topic Model (WNTM,~\cite{Zuo:2014}) and Biterm Topic Model (BTM,~\cite{Yan:2013a}) that use word co-occurrence data for analyzing short and imbalances texts. However, they do not consider their models as a way to learn word representations. There are also a number of papers on building hybrids of topic models and word embeddings. Gaussian LDA~\cite{Das:2015} imposes Gaussian priors for topics in a semantic vector space produced by word embeddings.
The learning procedure is obtained via Bayesian inference, however a similar idea is implemented more straightforwardly in~\cite{Sridhar:2015}. They use pre-built word vectors to perform clustering via Gaussian Mixture Model and apply the model to Twitter analysis. Pre-built word embeddings are also used in~\cite{Nguyen:2015} to improve quality of topic models on small or inconsistent datasets. Another model, called Topical Word Embeddings (TWE, ~\cite{Liu:2015}) combines LDA and SGNS. It infers a topic for each word occurrence and learns different embeddings for the same word occurred under different topics. Unlike all these models, we do not combine the models as separate mechanisms, but highlight a striking similarity of optimization objectives and \emph{merge} the models.

The rest of the paper is organized as follows. In section 2 we remind the basics of word embeddings and topic models. In sections 3 and 4 we discuss theoretic insights and introduce our generalized approach. In the experiments section we use 3 text datasets (Wikipedia, ArXiv, and Lenta.ru news corpus) to demonstrate high quality on word similarity and document similarity tasks, drastic improvement of interpretability and sparsity, and meaningful inter-modality similarities.





\section{Related work}

\paragraph{Definitions and notation.} Here we introduce the notation that highlights a common nature of all methods and will be used throughout the paper. Consider a set of documents~$D$ with a vocabulary~$W$. Let $n_{wd}$ denote a number of times the word~$w$ occurs in the document~$d$. The document can be treated as a \textit{global context}. 
We will be also interested in a \emph{local context} of each word occurrence, which is a bag of words in a window of a fixed size. Let $n_{uv}$ denote a number of co-occurrences of words~$u$ and $v$ in a sliding window, $n_u = \sum_v n_{uv}$, $n_v = \sum_u n_{uv}$, and $n = \sum_{u} n_u$.

All the models will be parametrized with the matrices $\Phi$ and $\Theta$, containing $|T|$-dimensional embeddings.

\paragraph{Skip-Gram model.}

Skip-gram model learns word embeddings by predicting a local context for each word in a corpus. 
The probability of word~$u$ from a local context of word~$v$ is modeled as follows:
\begin{equation}
\label{SGNS-puv}
    p(u|v) =
    \frac{\exp \sum_{t} \phi_{ut} \theta_{tv}}
    {\sum_{w \in W} \exp \sum_{t} \phi_{wt} \theta_{tv}},
\end{equation}
where $\Phi^{|W| \times |T|} = (\phi_{ut})$ and $\Theta^{|T| \times |W|} = (\theta_{tv})$ are two real-valued matrices of parameters. 
According to the bag-of-words assumption, each word in the local context is modeled independently, thus one can derive the log-likelihood as follows:

\begin{equation}
\label{SGNS-L}
\mathcal{L} = \sum_{v \in W} \sum_{u \in W} n_{uv} \ln p(u|v) \to \max\limits_{\Phi, \Theta}.
\end{equation}
where~$n_{uv}$ denotes the number of times the two terms co-occurred in a sliding window.
However, normalization over the whole vocabulary in formula~\eqref{SGNS-puv} prevents from learning the model effectively on large corpora. Skip-Gram Negative Sampling (SGNS) is one of possible ways to tackle this problem.  Instead of modeling a conditional probability~$p(u|v)$, SGNS models the probability of a co-occurrence for a pair of words~$(u, v)$. The model is trained on word pairs from the corpus (positive examples) as well as randomly sampled pairs (negative examples):
\begin{multline}
\label{SGNS:criteria}
    \sum_{v \in W} \sum_{u \in W} n_{uv} \log \sigma \left( \sum_{t} \phi_{ut} \theta_{tv} \right) +
    k\, \mathbb{E}_{\bar{v}} \log \sigma \left(- \sum_{t} \phi_{ut} \theta_{tv} \right) \to \max_{\Phi, \Theta},
\end{multline}
where~$\sigma$ is a sigmoid function, $\bar{v}$ are  sampled from unigram distribution and~$k$ is a parameter to balance positive and negative examples. SGNS model can be effectively learned via Stochastic Gradient Descent.

SGNS model can be extended to learn document representations if the probabilities in~\eqref{SGNS-puv} are conditioned on a document instead of a word. This architecture is called DBOW~\cite{Dai:2015} and it is one of the modifications of the popular paragraph2vec approach.

\paragraph{Topic model.}

Probabilistic Latent Semantic Analysis, PLSA~\cite{Hofmann:1999} is a topic model that describes words in documents by a mixture of hidden topics:

\begin{equation}
p(w|d) = \sum_{t \in T} p(w|t) p(t|d) = \sum_{t \in T} \phi_{wt} \theta_{td},
\end{equation}
where~$\Phi^{|W|\times |T|}$ contains probabilities~$\phi_{wt}$ of words in topics and $\Theta^{|T|\times |D|}$ contains probabilities~$\theta_{td}$ of topics in documents. 
The distributions are learned via maximization of the likelihood given normalization and non-negativity constraints:
\begin{align}
\label{PLSA-L}
& \mathcal{L} = \sum_{d \in D} \sum_{w \in W} n_{wd} \log p(w|d) \to \max_{\Phi,\,\Theta} \\
& \phi_{wt} \geq 0, \quad \sum_w \phi_{wt} = 1 \\
& \theta_{td} \geq 0, \quad \sum_t \theta_{td} = 1.
\end{align}
This task can be effectively solved via EM-algorithm~\cite{Deerwester:1990}
or its online modification~\cite{Hoffman:2010}. 
The most popular Latent Dirichlet Allocation \cite{Blei:2003} topic model extends PLSA by using Dirichlet priors for $\Phi$ and $\Theta$ distributions.

Additive Regularization of Topic Models, ARTM~\cite{Vorontsov:2015a} is a non-Bayesian framework for learning multiobjective topic models. The optimization task~\eqref{PLSA-L} is extended with $n$ additive regularizers $R_i(\Phi, \Theta)$ that are balanced with $\tau_i$ coefficients:
\begin{equation}
\label{ARTM-L}
\mathcal{L} + R \to \max_{\Phi,\,\Theta}; \quad
R = \sum_{i=1}^n \tau_i R_i(\Phi, \Theta)
\end{equation}
This approach addresses the problem of the non-uniqueness of the likelihood maximization~\eqref{PLSA-L} solution and imposes additional criteria to choose $\Phi$ and $\Theta$. The optimization is still done with online EM-algorithm, where M-step is modified to use the derivatives of the regularization terms~\cite{Vorontsov:2015a}.

\section{Probabilistic word embeddings}

Consider a modification of PLSA to predict the word~$u$ in a local context of the word~$v$: 
\begin{equation}
    p(u|v) = \sum_{t \in T} p(u|t) p(t|v) = \sum_{t \in T} \phi_{ut} \theta_{tv}
    \label{WW-PLSA-puv}
\end{equation}
In this formulation the topic model approximates a word co-occurrence matrix instead of a word-document matrix. Unlike in PLSA, $\Theta^{|T|\times |W|}$ contains probabilities~$\theta_{tv}$ of topics for \textit{words}. However, from the topic modeling perspective, those words can be treated as  \textit{pseudo-documents}. One may think of a pseudo-document \textit{derived by a word~$v$} as a concatenation of all local contexts for all occurrences of the word~$v$ in the corpus. 
A local context is still defined as a fixed-size window, but this definition can be easily extended to use syntactic patterns, sentences, or any other structure. 

Interestingly, this approach appears to be extremely similar to Skip-Gram model~\eqref{SGNS-puv}. Both models predict the same probabilities $p(u|v)$ and make use of the observed data by optimizing exactly the same likelihood~\eqref{SGNS-L}. Both models are parametrized with matrices of hidden representations of words. The only difference is the space of the parameters: while Skip-Gram has no constraints, the topic model learns non-negative and normalized vectors that have a probabilistic interpretation. As a benefit, word probabilities can be predicted with a mixture model of the parameters with no need in explicit \textit{softmax} normalization.


Learning probabilistic word embeddings (PWE) can be treated as a stochastic matrix factorization of probabilities $p(u|v)$ estimated from a corpus.
This makes a perfect analogy with matrix factorization formulations of SGNS~\cite{Levy:2014a}, GloVe, NNSE, and other similar techniques. GloVe uses a squared loss with a weighting function~$f(n_{uv})$ that penalizes too frequent co-occurrences. Apart from two real-valued matrices of parameters, it introduces  bias terms~$b_u$ and~$\tilde{b}_v$. NNSE also uses a squared loss, but imposes additional constraints to obtain sparse non-negative embeddings~$\phi_u$ and guarantees the limited $l2$-norm for $\Theta$ rows, which are called \emph{dictionary} entries.

We summarize the connections between all mentioned models in Table \ref{table_mf}. Each method is decomposed into several components: the type of raw  co-occurrence data $F=(f_{uv})^{W \times W}$, the matrix factorization loss, the constraints for a parameter space, and the optimization technique.
From this point of view, there is no big difference between so called \emph{count-based} and \emph{predictive} approaches. On the one hand, each method counts $f_{uv}$ values (probably implicitly) and performs dimensionality reduction by a matrix factorization. On the other hand, each matrix factorization objective can be treated as a loss, which is used to train the model from data. 
More importantly, the unified view provides a powerful tool to analyze a diverse set of existing models and exchange components across them.

\begin{table*}[t]
\centering
\caption{Learning word embeddings with a low-rank matrix factorization.}
\label{table_mf}
\begin{tabular}{|c|c|c|}
\hline

PWE & 

\begin{tabular}{c} 
data type \\ \hline
objective \\ \hline
constrains \\ \hline
technique \\ 
\end{tabular} & 

\begin{tabular}{c} 
$F_{uv} = \frac{n_{uv}}{n_v} = \hat{p}(u|v)$ \\ \hline

$ \sum_{v \in W} n_v \KL\left(\hat{p}(u|v) \big|\big| \left< \phi_{u} \theta_{v} \right>\right) \to \min\limits_{\Phi, \Theta} $ \\ \hline

$ \phi_{ut} > 0, \quad \sum_{u} \phi_{ut} = 1; \quad
\theta_{tv} > 0, \quad \sum_{t} \theta_{tv} = 1$ \\ \hline

EM-algorithm (online by $F$ columns) \\ 
\end{tabular} \\ \hline

SGNS & 

\begin{tabular}{c} 
data type \\ \hline
objective \\ \hline
constrains \\ \hline
technique \\ 
\end{tabular} & 

\begin{tabular}{c} 
$F_{uv} = \log \frac{n_{uv}n}{n_u n_v} - \log k$ \\ \hline

$\sum_{u \in W} \sum_{v \in W} n_{uv} \log \sigma \left( \left< \phi_{u} \theta_{v}\right> \right) +
k\, \mathbb{E}_{\bar{v}} \log \sigma \left(- \left< \phi_{u} \theta_{v}\right> \right) \to \max_{\Phi, \Theta}$ \\ \hline

No constraints \\ \hline

SGD (online by corpus) \\ 
\end{tabular} \\ \hline

GloVe & 

\begin{tabular}{c} 
data type \\ \hline
objective \\ \hline
constrains \\ \hline
technique \\ 
\end{tabular} & 

\begin{tabular}{c} 
$F_{uv} = \log n_{uv}$ \\ \hline

$\sum_{v \in W} \sum_{u \in W} f(n_{uv}) \big( \left<  \phi_{u} \theta_{v} \right> + b_u + \tilde{b}_v -
     \log n_{uv} \big)^2 
     \to \min_{\Phi, \Theta, b, \tilde{b}}$ \\ \hline
No constraints \\ \hline
AdaGrad (online by $F$ elements) \\ 
\end{tabular} \\ \hline

NNSE & 
\begin{tabular}{c} 
data type \\ \hline
objective \\ \hline
constrains \\ \hline
technique \\ 
\end{tabular} & 

\begin{tabular}{c} 
$F_{uv} = max(0, \log \frac{n_{uv}n}{n_u n_v})$ 
or SVD low-rank approximation\\ \hline

$\sum_{u \in W} \left(\| f_u - \phi_u \Theta \|^2 + \| \phi_u \|_{1} \right) \to\min_{\Phi, \Theta}$ \\ \hline
 
$\phi_{ut} \geq 0, \forall u \in W, t \in T \quad 
\theta_t \theta_t^T \leq 1, \forall t \in T$
\\ \hline

Online algorithm from~\cite{Mairal:2010} \\

\end{tabular} \\ \hline

\end{tabular}
\end{table*}

\section{Additive regularization and embeddings for multiple modalities}

The proposed probabilistic embeddings can be easily extended as a topic model. First, there is a natural way to learn document embeddings. 
Second, additive regularization of topic models~\cite{Vorontsov:2015a} can be used to meet further requirements.
In this paper we employ it to obtain a high sparsity with no reduction in the accuracy of matrix factorization.
The regularization criteria is a sum of cross-entropy terms between the target and fixed distributions:
\begin{equation}
R = - \tau \sum_{t \in T}\sum_{u \in W} \beta_u \ln \phi_{ut} 
\end{equation}
where~$\beta_u$ can be set to the uniform distribution.

Furthermore, we extend the topic model to incorporate meta-data or \textit{modalites}, such as timestamps, categories, authors, etc. 
Real data often has such type of information associated with each document and it is desirable to build representations for these additional tokens as well as for the usual words.

Recall that each \textit{pseudo-document}~$v$ in our training data is formed by collecting words $u$ that co-occur with word $v$ within a sliding window. Now we enrich it by the tokens $u$ of some additional modality $m$ that co-occur with the word $v$ within a document. 
The only difference here is in using \textit{global} document-based co-occurrences for additional modalities as opposed to \textit{local} window-based co-occurrences for the modality of words. 
Once the \textit{pseudo-documents} are prepared, we employ Multi-ARTM approach~\cite{Vorontsov:2015b} to learn topic vectors for tokens of each modality:
 \begin{align}
 \label{Multi-ARTM}
            &\sum_{m\in M} \lambda_m
            \underbrace{
                \sum_{v\in W^0} \sum_{u\in W^m} \!n_{uv} \ln p(u|v)
            }_{\text{modality log-likelihood~} \mathcal{L}_m(\Phi,\Theta)}
            \to \max_{\Phi,\Theta},\\
            &\phi_{ut}\geq 0,~
            \sum_{u\in W^m}\!\! \phi_{ut}=1, \,
            \forall m\in M;
            \\
            &\theta_{tv}\geq 0,\,\,\,\,\,~
            \sum_{t\in T} \theta_{tv}=1.
        \end{align}
where $\lambda_m>0$ are \emph{modality weights},
$W^m$ are modality vocabularies, and $m=0$ for the basic text modality. Optionally, the tokens of other modalities can also form pseudo-documents and this would restore the symmetric property of the factorized matrix. Regularizers can be still added to the multimodal optimization criteria. 
        

\paragraph{Online EM-algorithm.}
Regularized multimodal likelihood maximization is performed with online EM-algorithm implemented in BigARTM library~\cite{Frei:2015}.
First, we compute all necessary co-occurrences and build the \textit{pseudo-documents} as described before. We store this corpus on disk and process it by batches of $B=100$ pseudo-documents.
The algorithm starts with random initialization of $\Phi$ and $\Theta$ matrices.
The E-step estimates posterior topic distributions~$p(t|u, v)$ for words~$u$ in a pseudo-document~$v$. These updates are alternating with $\theta_{tv}$ updates for the given pseudo-document. After a fixed number of iterations through the pseudo-document, $\theta_{tv}$ are thrown away, while $p(t|u,v)$ are used to compute incremental unnormalized updates for $\phi_{ut}$. These updates are applied altogether when the whole batch of pseudo-documents is processed. Importantly, these procedure does not overwrite the previous value of $\Phi$, but slowly forgets it with an exponential moving average. 
The detailed formulas for the case of usual documents can be found in~\cite{Frei:2015}.
Note that the only matrix which has to be always stored in RAM is $\Phi$. The number of epochs (runs through the whole corpus) in our experiments ranges from 1 to 6.

\section{Experiments}
 
\begin{table*}[t]
\centering
\caption{Spearman correlation for word similarities on Wikipedia.}
\label{table1}
\begin{tabular}{|c|c|c|c|c|c|c|c|c|}
\hline
$\,$ Model $\,$ & $\,$ Data $\,$ & Optimization & Metric &
\begin{tabular}[c]{@{}c@{}}WordSim \\ Sim.\end{tabular} & \begin{tabular}[c]{@{}c@{}}WordSim \\ Rel.\end{tabular} & WordSim        & \begin{tabular}[c]{@{}c@{}}Bruni MEN\end{tabular} & 
SimLex-999 \\ \hline

LDA & $n_{wd}$ & online EM & hel &  0.530 & 0.455 & 0.474 & 0.583 & 0.220 \\ \hline 

PWE & $n_{uv}$ &  offline EM  & dot & 0.709	& 0.635	& 0.654	& 0.658	& 0.240 \\ \hline


PWE & pPMI & offline EM & dot & 0.701 &	0.615	& 0.647 &	0.707 &	 0.276 \\ \hline
PWE & $n_{uv}$ & online EM & dot & 0.718 &	\textbf{0.673} & \textbf{0.685} &	0.669 & 0.263 \\ \hline

SGNS & sPMI & SGD & cos & \textbf{0.752} &	0.632	& 0.666	& \textbf{0.745} & \textbf{0.384} \\ \hline

\end{tabular}
\end{table*}

We conduct experiments on three different datasets. Firstly, we compare the proposed Probabilistic Word Embeddings (PWE) to SGNS on Wikipedia dump by word similarities and interpretability of the components. Secondly, we learn probabilistic document embeddings on ArXiv papers and compare them to DBOW on the document similarity task~\cite{Dai:2015}. Finally, we learn embeddings for multiple modalities on a corpus of Russian news Lenta.ru and investigate inter-modality similarities. 
All topic models are learnt in BigARTM\footnote{bigartm.org} open source library~\cite{Frei:2015} using Python interface\footnote{github.com/bigartm/bigartm-book/blob/master/applications/word\_embeddings.ipynb}. 
SGNS is taken from Hyperwords\footnote{bitbucket.org/omerlevy/hyperwords} package and DBOW is taken from Gensim\footnote{radimrehurek.com/gensim/} library.

\begin{figure}[t]
\centering
\begin{minipage}{0.49\textwidth}
\includegraphics[width=0.9\textwidth]{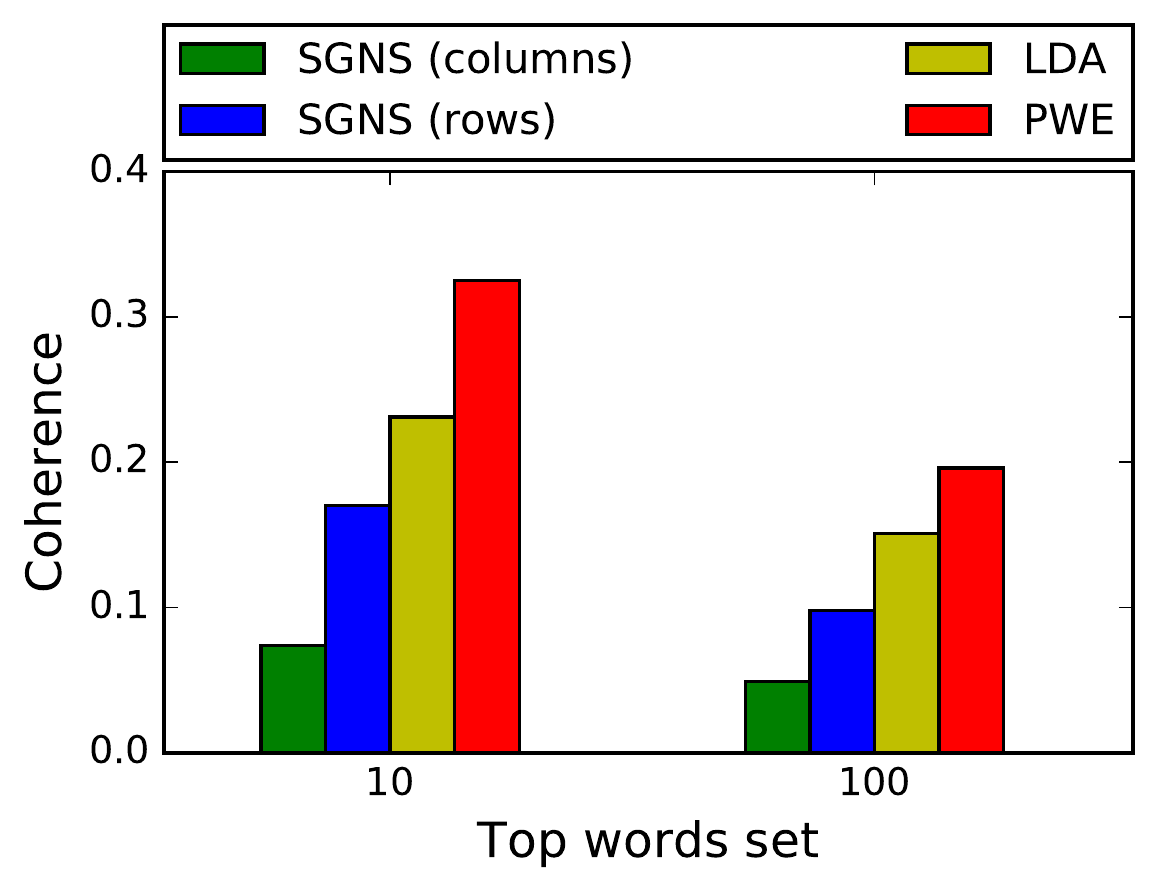}
\caption{Coherence scores.}
\label{coherencefigure}
\end{minipage}%
\begin{minipage}{0.49\textwidth}
\includegraphics[width=0.93\textwidth]{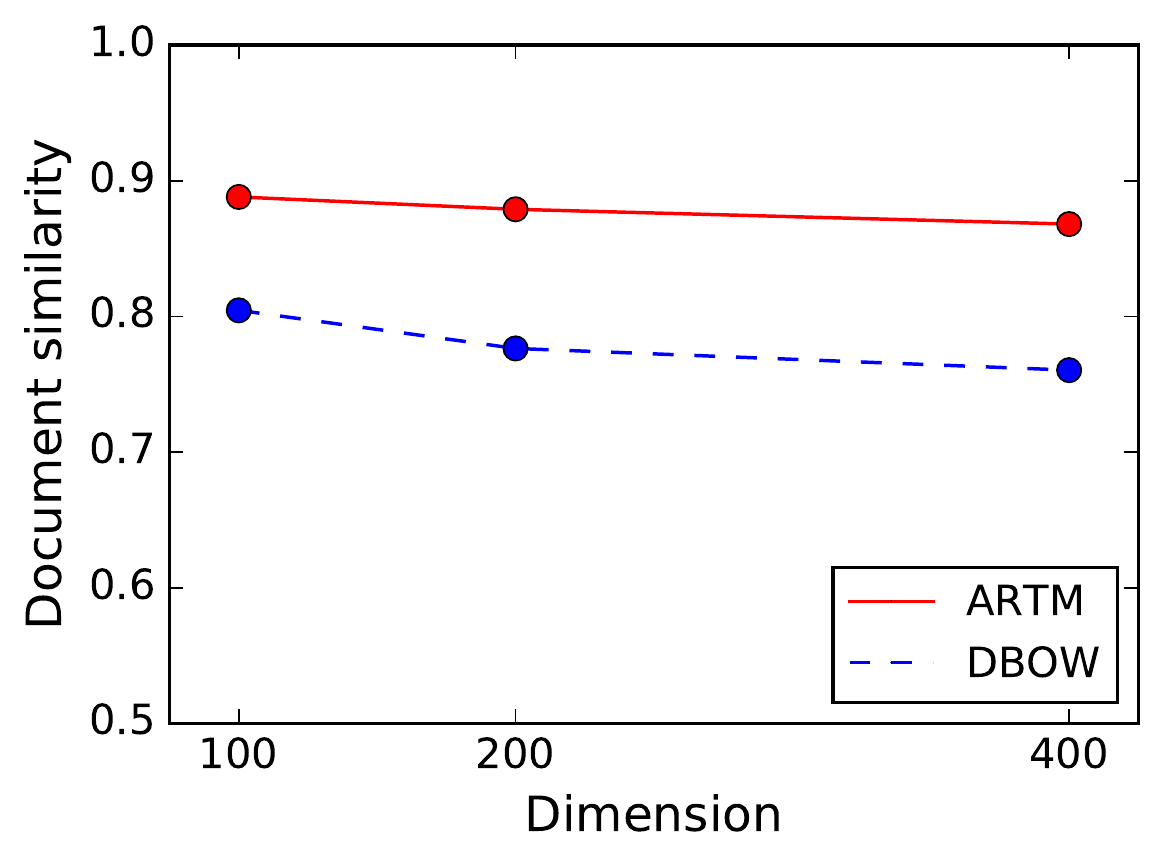}
\caption{Document similarities.}
\label{docsimfigure}
\end{minipage}
\end{figure}

\paragraph{Word similarity tasks.}

We use Wikipedia 2016-01-13 dump and preprocess it with Levy's scripts$^2$ to guarantee equal conditions for SGNS and topic modeling~\cite{Levy:2015}.
We delete top $25$ stop-words from the vocabulary, keep the next $100000$ words, and delete the word pairs that co-occur less than $5$ times.
We performed experiments for \textit{windows of size $2$, $5$, and $10$}, but report here only window-$5$ results, as the others are analogous. 
We use \textit{subsampling} with the constant $10^{-5}$ for all models.
While common for SGNS, subsampling has never been used for topic modeling. However, our experiments show that it slightly improves topic interpretability by filtering out too general terms and therefore might be a good preprocessing recommendation.
Also, we tried using \textit{dynamic} window, which is a weighting technique based on the distance of the co-occurred words, but we didn't find it beneficial. 

Following a traditional benchmark for word similarity tasks, we rank word pairs according to our models and measure Spearman correlation with the human ratings from WordSim353 dataset~\cite{Finkelstein:2002} partitioned into  WordSim Similarity and WordSim Relatedness~\cite{Agirre:2009}, MEN dataset~\cite{Bruni:2012}, and SimLex-999~\cite{Hill:2015}.
We consider SGNS model as a baseline and investigate if probabilistic word embeddings (PWE) are capable of providing the comparable quality. 
We start with LDA and Hellinger distance for word vectors as this is the default choice from many papers, e.g.~\cite{Murphy:2012}. Table~\ref{table1} shows that SGNS dramatically outperforms LDA. Our further experiments demonstrate how to make topic models work. 

First, we get an improvement by modeling the word-word matrix instead of the word-document matrix. Second, we investigate how to compute word similarity in the obtained space of probabilistic embeddings. We find the topic distributions should be normalized using Bayes' rule $p(t|u) = \frac{\phi_{ut} p(t)}{\sum_{t} \phi_{ut} p(t)}$ and that dot-product performs better than Hellinger distance or cosine similarity. Third, we find that online EM-algorithm with incremental $\Phi$ updates performs better than its offline analogue, where $\Phi$ is overwritten once per epoch. We also find that it is beneficial to initialize $\Theta$ randomly each time rather than store the values from the previous epoch.
This combination of tricks gives the accuracy comparable to SGNS. 

To obtain \textit{sparsity}, we add the regularizer at the last iterations of EM-algorithm and observe \textbf{93\%} of zeros in word embeddings \textit{with the same} performance on word similarity tasks. 
We also try different co-occurrence scores instead of raw counts such as $\log n_{uv}$ to penalize frequent co-occurrences or normalized $\frac{n_{uv}}{\sum_u n_{uv}}$ values to obtain a sum of \textit{non-weighted} KL-divergences in the optimization criteria. While most of these weighting schemes give worse results, positive PMI values appear to be beneficial for some testsets.

\paragraph{Interpretability of embedding components.}

\begin{table}[t]
\begin{minipage}{0.53\textwidth}
\centering
\caption{Interpretability of topics.}
\label{coherence-table}
\begin{tabular}{|cc|cc|}
\hline
\multicolumn{2}{|c|}{\begin{tabular}[c]{@{}c@{}}PWE\\ { }  \end{tabular}}    & \multicolumn{2}{|c|}{\begin{tabular}[c]{@{}c@{}}SGNS\\ { } \end{tabular} } \\ \hline
art        & arbitration          & transports   & rana     \\
painting   & ban                  & recon     & walnut \\
museum     & requests             & grumman       & rashid     \\
painters    & arbitrators            & convoys     & malek     \\
gallery    & noticeboard          & piloted       & aziz     \\
sculpture  & block                & stealth   & khalid     \\
painter   & administrators     & flotilla         & yemeni      \\ 
exhibition & arbcom          & convoy          & andalusian     \\
portraits   & sanctions          & supersonic    & bien     \\
drawings   & mediation            & bomber        & gcc     \\ \hline

\end{tabular}
\end{minipage}%
\begin{minipage}{0.45\textwidth}
\centering
\caption{Event timestamps.}
\label{lenta_examples}
\begin{tabular}{|c|c|c|}
\hline
\begin{tabular}[c]{@{}c@{}}2015-12-18\\ SW release \end{tabular} & \begin{tabular}[c]{@{}c@{}}2016-02-29\\ The Oscars\end{tabular} & \begin{tabular}[c]{@{}c@{}}2015-05-09\\ Victory Day\end{tabular} \\ \hline
jedi & statuette & great   \\
sith  & award   & anniversary      \\
fett   & nomination   & normandy  \\
anakin   & linklater    & parade    \\
chewbacca    &  oscar     & demonstration  \\
film series &  birdman  &  vladimir \\
hamill & win &  celebration \\
prequel  & criticism    & concentration    \\
awaken & director & auschwitz \\
boyega   & lubezki  &  photograph \\ \hline
\end{tabular}
\end{minipage}
\end{table}

We characterize each component by a set of words with the highest values in the embedding matrix and check if those sets correspond to some aspects that can be named by a human.
\textit{Word intrusion}~\cite{Boyd-Graber:2009} technique
is based on the idea that for well formed sets, a human expert can easily detect an intruder, randomly sampled from the vocabulary. 
This technique has been widely used in topic modeling and also for Non-Negative Sparse Embeddings~\cite{Murphy:2012} and Online Interpretable Word Embeddings~\cite{Luo:2015}. 
Word intrusion requires experts, but it can be automated by the \textit{coherence score}, which is shown to have high correlations with human judgements~\cite{Newman:2010}. It averages pairwise similarities across the set of words. For  similarities one can use PMI scores from an external corpus~\cite{Newman:2011}, log-conditional probabilities from the same corpus~\cite{Mimno:2011}, distributional similarities~\cite{Aletras:2013}, or other variants~\cite{Roder:2015}.

In our experiments we use the PMI-based coherence for top-$10$ and top-$100$ words for each component. The score is averaged over the components and reported in Figure~\ref{coherencefigure}. 
For SGNS we consider two different schemes of ranking words within each component. First, using the raw values; second, applying softmax \textit{by rows} and using Bayes' rule to convert $p(t|w)$ into $p(w|t)$ probabilities. We show that the coherence for probabilistic word embeddings is consistently higher than that of LDA or SGNS for a range of embedding sizes. Also, this result is confirmed by visual analysis of the obtained components (see Table~\ref{coherence-table} for the examples).

\begin{table*}[t]
\centering
\caption{Spearman correlation for word similarities on Lenta.ru.}
\label{lenta_sim}
\begin{tabular}{|c|c|c|c|c|c|c|}
\hline
Model  & WordSim Sim & WordSim Rel &  $\quad$ MC $\quad$ &  $\quad$ RG  $\quad$ &   $\quad$ HJ  $\quad$ &  $\,\,$ SimLex $\,\,$ \\ \hline

SGNS &  0.630 & 0.530 &	0.377 &	0.415 &	0.567 & \textbf{0.243} \\ 
\hline
CBOW &  0.625 &	0.513 &  0.403 & 0.370	& 0.551 & 0.170 \\ 
\hline
PWE &  0.649	&  0.565 &	 0.605 & \textbf{0.594} & 0.604 & 0.123\\ \hline 
Multi-PWE &  \textbf{0.682} & \textbf{0.58} &	\textbf{0.607} &  0.584 &	 \textbf{0.611} & 0.144 \\ \hline

\end{tabular}
\end{table*}

\paragraph{Document similarity task.}

In this experiment we learn probabilistic document embeddings on ArXiv corpus and test them on a document similarity task. The testset released by Dai et. al~\cite{Dai:2015} contains automatically generated triplets of a query paper, a similar paper that shares  key words, and a dis-similar paper that does not share any key words. The quality is evaluated by the accuracy of identifying the similar one within each triplet.
We preprocess\footnote{https://github.com/romovpa/arxiv-dataset} plain texts of $963564$ ArXiv papers with a total of~$1416554733$ tokens   
and reduce the vocabulary size to $122596$ words with a frequency-based filtering. The restored mapping between the plain texts and the URLs from the testset\footnote{http://cs.stanford.edu/˜quocle/triplets-data.tar.gz} covers $15853$ triplets out of $20000$. 

We train embeddings with $1$ epoch of online EM-algorithm. Note that the matrix $\Theta$ is not stored, so memory consumption does not grow linearly with the number of documents. Afterwards, we infer test embeddings with $10$ passes on each document. 
As a baseline, we train DBOW~\cite{Dai:2015} with $15$ epochs and use linear decay of learning rate from $0.025$ to $0.001$; afterwards we infer test embeddings with $5$ epochs. Unlike online EM-algorithm, DBOW needs in-memory storage of document vectors and also takes much longer to train (several hours instead of $30$ minutes on the same machine).
We do not facilitate training word vectors in DBOW, because it slows down the process dramatically.

Figure \ref{docsimfigure} shows that our ARTM model consistently outperforms DBOW for a range of embedding sizes. The absolute numbers are also better than for all other methods reported in~\cite{Dai:2015}, thus giving a new state-of-the-art on this dataset.

\paragraph{Multimodal embedding similarities.}

The experiments are held on Russian \textit{lenta.ru} corpus, that contains $100033$ news with a total of $10050714$ tokens. 
The corpus has additional modalities of timestamps ($825$ unique tokens), categories ($22$ unique tokens) and sub-categories ($97$ unique tokens). The basic text modality has $54963$ unique words.

We produce a collection of pseudo-documents using the window of size 5 and subsampling. For evaluation we use HJ testset~\cite{panchenko:2016} with human judgments on 398 word pairs translated to Russian from the widely used English testsets: MC \cite{Miller:1991}, RG~\cite{Rubenstein:1965}, and WordSim353~\cite{Finkelstein:2002}. We also use SimLex-999 testset translation~\cite{Leviant:2015}.

Table \ref{lenta_sim} shows that probabilistic word embeddings (PWE)  outperform SGNS for most of the testsets even without using additional modalities. 
One can note that this corpus is relatively small and it might be a reason for poor SGNS performance. We have also tried CBOW~\cite{Mikolov:2013} following a common recommendation to use it for small data, but it performed even worse. Generally, we observe that topic modeling requires less data for a good performance, thus the proposed PWE approach might be beneficial for applications with limited data.

Next, we use additional modalities and optimize the modality weights in the objective~\eqref{Multi-ARTM}. With this approach we observe a further boost in the performance for the word similarity task (see Multi-PWE in Table \ref{lenta_sim}). 
Finally, we experiment with two different modes: using modalities only as tokens (a non-symmetric case) and both as tokens and pseudo-documents (a symmetric case).  While word similarities are better for the non-symmetric case, we observe better inter-modality similarities for the symmetric case. Table \ref{lenta_examples} provides several examples of remarkable timestamps and their closest words. The words are manually translated from Russian to English for reporting purposes only. Each column is easily interpretable as a coherent event, namely the release of Star Wars, the Oscars 2016, and Victory Day in Russia.

\section{Conclusions}

In this work we revisited topic modelling techniques in the context of learning hidden representations for words and documents. Topic models are known to provide interpretable components but perform poorly on word similarity tasks.
However, we have shown that topic models and neural word embeddings can be made to predict the same probabilities with the only difference in the probabilistic nature of parameters. This theoretical insight enabled us to merge the models and get practical results. First, we obtained probabilistic word embeddings (PWE) that work on par with SGNS on word similarity tasks, but have high sparsity and interpretability of the components. Second, we learned document embeddings that outperform DBOW on a document similarity task and require less memory and time for training. Furthermore, considering the task as a topic modeling, enabled us to adapt Multi-ARTM approach and learn embeddings for multiple modalities, such as timestamps and categories. We observed meaningful inter-modality similarities and a boost of the quality on the basic word similarity task.
In future we plan to apply the proposed probabilistic embeddings to a suite of NLP tasks and take even more advantage of the additive regularization to incorporate task-specific requirements into the models. 

\subsubsection*{Acknowledgements.}
The work was supported by~Government of the Russian Federation (agreement 05.Y09.21.0018) 
and the Russian Foundation for Basic Research grants 17-07-01536, 16-37-00498.

\bibliography{ainl2017}
\bibliographystyle{splncs03}

\end{document}